\documentclass{ifacconf}%

\usepackage{amsfonts}
\usepackage{amsmath}
\usepackage{comment}
\usepackage{graphicx}
\usepackage{amssymb,epsfig}
\usepackage[authoryear]{natbib}
\usepackage{amssymb}
\usepackage{algorithm}
\usepackage{algorithmic}

\begin{document}

\begin{frontmatter}

\title{An Online Expectation-Maximisation Algorithm for Nonnegative Matrix Factorisation Models}

\author[1]{Sinan Y{\i}ld{\i}r{\i}m},
\author[2]{A. Taylan Cemgil},
\author[3]{Sumeetpal S. Singh}
\address[1]{Statistical Laboratory, DPMMS, University of Cambridge, UK}
\address[2]{Department of Computer Engineering, Bo\v{g}azi\c{c}i University, Turkey}
\address[3]{Department of Engineering, University of Cambridge, UK}
          

\begin{abstract}                         
In this paper we formulate the nonnegative matrix factorisation (NMF) problem as a maximum likelihood estimation problem for hidden Markov models and propose online expectation-maximisation (EM) algorithms to estimate the NMF and the other unknown static parameters. We also propose a sequential Monte Carlo approximation of our online EM algorithm. We show the performance of the proposed method with two numerical examples.
\end{abstract}

\end{frontmatter}

\section{Introduction} \label{sec: Introduction}
With the advancement of sensor and storage technologies, and with the cost of data acquisition dropping significantly, we are able to collect and record vast amounts of raw data. Arguably, the grand challenge facing computation in the 21st century is the effective handling of such large data sets to extract meaningful information for scientific, financial, political or technological purposes \citep{Donoho2000}. Unfortunately, classical batch processing methods are unable to deal with very large data sets due to memory restrictions and slow computational time.

One key approach for the analysis of large datasets is based on the matrix and tensor factorisation paradigm. Given an observed dataset $Y$, where $Y$ is a matrix of a certain dimension and each element of it corresponds to an observed data point, the matrix factorisation problem is the computation of matrix factors $B$ and $X$ such that $Y$ is approximated by the matrix product $BX$, i.e.,
\begin{equation} \label{eq:matfact}
Y \approx B X \nonumber
\end{equation}
(Later we will make our notation and inferential goals more precise.) Indeed, many standard statistical  methods such as clustering, independent components analysis (ICA), nonnegative matrix factorisation (NMF), latent semantic indexing (LSI), collaborative filtering can be expressed and understood as matrix factorisation problems \citep{Lee_and_Seung_1999, singh08, Koren2009}. 

Matrix factorisation models also have well understood probabilistic/statistical interpretations as probabilistic generative models and many standard algorithms mentioned above can also be derived as maximum likelihood or maximum a-posteriori parameter estimation procedures \citep{cedric-eusipco2009, salakhut08, cemgil09}. The advantage of this interpretation is that it enables one to incorporate domain specific prior knowledge in a principled and consistent way. This can be achieved by building hierarchical statistical models to fit the specifics of the application at hand. Moreover, the probabilistic/statistical approach also provides a natural framework for sequential processing which is desirable for developing online algorithms that pass over each data point only once.  While the development of effective online algorithms for matrix factorisation are of interest on their own, the algorithmic ideas can be generalised to more structured models such as tensor factorisations (e.g. see \citep{Kolda2009}).

In this paper our primary interest is estimation of $B$ (rather than $B$ and $X$), which often is the main objective in NMF problems. We formulate the NMF problem as a maximum likelihood estimation (MLE) problem for hidden Markov models (HMMs). The advantage of doing so is that the asymptotic properties of MLE for HMM's has been studied in the past by many authors and these results may be adapted to the NMF framework. We propose a sequential Monte Carlo (SMC) based online EM algorithm \citep{Cappe_2009a, Del_Moral_et_al_2009} for the NMF problem. SMC introduces a layer of bias which decreases as the number of particles in the SMC approximation is increased.

In the literature, several online algorithms have been proposed for online computation of matrix factorisations. \citet{marial09} propose an online optimisation algorithm, based on stochastic approximations, which scales up gracefully to large data sets with millions of training samples. A proof of convergence is presented for the Gaussian case. There are similar formulations applied to other matrix factorisation formulations, notably NMF \citep{lefevre2011} and Latent Dirichlet Allocation \citep{hoffman2010}, as well as alternative views for NMF which are  based on incremental subspace learning \citep{bucak09}. Although the empirical results of these methods suggest good performance, their asymptotic properties have not been established.

\subsection{Notation}
Let $A$ be a $M \times N$ matrix. The $(m, n)$'th element of $A$ is $A(m, n)$. If $M$ (or $N$) is 1, then $A(i) = A(1, i)$ (or $A(i, 1)$). The $m$'th row of $A$ is $A(m, \cdot)$. If $A$ and $B$ are both $M \times N$ matrices, $C = A \odot B$ denotes element-by-element multiplication, i.e., $C(m, n) = A(m, n)B(m, n)$; $\frac{A}{B}$ (or $A / B$) means element-by-element division, in a similar way. $\mathbf{1}_{M \times N}$ ($\mathbf{0}_{M \times N}$) is a $M \times N$ matrix of $1$'s ($0$'s), where $\mathbf{1}_{M \times 1}$ is abbreviated to $\mathbf{1}_{M}$. $\mathbb{N} = \{ 0, 1, 2, \ldots \}$ and $\mathbb{R}_{+} = \left[ 0, \infty \right)$ are the sets of nonnegative integers and real numbers. Random variables will be defined by using capital letters, such as $X, Y, Z$, etc.,  and their realisations will be corresponding small case letters ($x, y, z$, etc.). The indicator function $I_{\alpha}(x)  = 1$ if $x = \alpha$, otherwise it is $0$; also, for a set $A$, $I_{A}(x) = 1$ if $x \in A$, otherwise it is $0$.

\section{The Statistical Model for NMF} \label{sec: The Statistical Model for NMF}
Consider the following HMM comprised of the latent processes $\left\{ X_{t}, Z_{t} \right\}_{t \geq 1}$ and the observation process $\left\{ Y_{t} \right\}_{t \geq 1}$. The process $\left\{ X_{t} \in \mathbb{R}_{+}^{K} \right\}_{t \geq 1}$ is a Markov process of $K \times 1$ nonnegative vectors with an initial density $\mu_{\psi}$ and the transition density $f_{\psi}$ for $t = 2, 3, \ldots$
\begin{eqnarray} \label{eq: law of X}
X_{1} \sim \mu_{\psi}(x), \text{ } X_{t}  | \left( X_{t-1} = x_{t-1} \right) \sim f_{\psi}(x_{t} | x_{t-1}),
\end{eqnarray}
where $\psi \in \Psi$ is a finite dimensional parameter which parametrizes the law of the Markov process. $Z_{t} \in \mathbb{N}^{M \times K}$ is a $M \times K$ matrix of nonnegative integers, and its elements are independent conditioned on $X_{t}$ as follows:
\begin{eqnarray} \label{eq: law of Z given X}
Z_{t} | \left( X_{t} = x_{t} \right) \sim \prod_{m = 1}^{M} \prod_{k = 1}^{K} \mathcal{PO}(z_{t}(m, k); B(m,k) x_{t}(k)) \nonumber
\end{eqnarray}
where $B \in \mathbb{R}_{+}^{M \times K}$ is an $M \times K$ nonnegative matrix. Here $\mathcal{PO}(v; \lambda)$ denotes the Poisson distribution on $\mathbb{N}$ with intensity parameter $\lambda \geq 0$
\begin{equation} \label{eq: Poisson distribution}
\mathcal{PO}(v; \lambda) = \exp \left( v \log \lambda - \lambda - \log v! \right) \nonumber,
\end{equation}
The $M \times 1$ observation vector $Y_{t}$ is conditioned on $Z_{t}$ in a deterministic way
\begin{equation} \label{eq: law of Y given Z}
Y_{t}(m) =  \sum_{k = 1}^{K}{Z_{t}(m, k)}, \quad m =1, \ldots, M. \nonumber
\end{equation}
This results in the conditional density of $Y_{t}$ given $X_{t} = x_{t}$, denoted by $g_{B}$, being a multivariate Poisson density
\begin{eqnarray} \label{eq: law of Y given X}
Y_{t} | \left( X_{t} = x_{t} \right) \sim g_{B}(y_{t} | x_{t}) = \prod_{m = 1}^{M} \mathcal{PO} \left( y_{t}(m) ; B(m, \cdot) x_{t} \right).
\end{eqnarray}
Hence the likelihood of $y_{t}$ given $x_{t}$ can analytically be evaluated. Moreover, the conditional posterior distribution $\pi_{B}(z_{t}| y_{t}, x_{t})$ of $Z_{t}$ given $y_{t}$ and $x_{t}$ has a factorized closed form expression:
\begin{eqnarray} \label{eq: law of Z given X and Y}
Z_{t} | \left( Y_{t} = y_{t}, X_{t} = x_{t} \right) & \sim & \pi_{B}(z_{t} | y_{t}, x_{t}) \nonumber \\
&=& \prod_{m = 1}^{M} \mathcal{M} \left( z_{t}(m, \cdot) ; y_{t}(m), \rho_{t, m} \right)
\end{eqnarray}
where $\rho_{t, m}(k) = B(m, k) x_{t}(k) /  B(m, \cdot ) x_{t}$ and $\mathcal{M}$ denotes a multinomial distribution defined by
\begin{eqnarray} \label{eq: multinomial distribution}
\mathcal{M}(v; \alpha, \rho) = I_{\alpha} \left( \sum_{k = 1}^{K} v_{k} \right) \alpha! \prod_{k = 1}^{K} \frac{\rho_{k}^{v_{k}}}{v_{k}!}, \nonumber
\end{eqnarray}
where $v = \left[ v_{1} \ldots v_{K} \right]$ is a realisation of the vector valued random variable $V = \left[ V_{1} \ldots V_{K} \right]$, $\rho = \left( \rho_{1}, \ldots, \rho_{K} \right)$, and $\sum_{k = 1}^{K} \rho_{k} = 1$. It is a standard result that the marginal mean of the $k$'th component is $\mathbb{E}_{\alpha, \rho}\left[ V_{k} \right] =  \alpha \rho_{k}$.

Let $\theta = \left( \psi, B \right) \in \Theta = \Psi \times \mathbb{R}_{+}^{M \times K}$ denote all the parameters of the HMM. We can write the joint density of $( X_{1:t}, Z_{1:t}, Y_{1:t}) $ given $\theta$ as
\begin{eqnarray} \label{eq: joint density of X, Y, and Z}
p_{\theta}(x_{1:t}, z_{1:t}, && y_{1:t}) = \mu_{\psi}(x_{1})g_{B}(y_{1}|x_{1})\pi_{B}(z_{1}|y_{1}, x_{1}) \nonumber \\
&&\times  \prod_{i = 2}^{t} f_{\psi}(x_{i}|x_{i-1}) g_{B}(y_{i} | x_{i}) \pi_{B}(z_{i} | x_{i}, y_{i}).
\end{eqnarray}
From \eqref{eq: joint density of X, Y, and Z}, we observe that the joint density of $(X_{1:t}, Y_{1:t})$
\begin{eqnarray} \label{eq: joint density of X and Y}
p_{\theta}(x_{1:t}, y_{1:t}) =  \mu_{\psi}(x_{1}) g_{B}(y_{1}|x_{1}) \prod_{i = 2}^{t} f_{\psi}(x_{i}|x_{i-1}) g_{B}(y_{i} | x_{i}) \nonumber
\end{eqnarray}
defines the law of another HMM $\left\{ X_{t}, Y_{t} \right\}_{t \geq 1}$ comprised of the latent process $\left\{ X_{t} \right\}_{t \geq 1}$, with initial and transitional densities $\mu_{\psi}$ and $f_{\psi}$, and the observation process $\left\{ Y_{t} \right\}_{t \geq 1}$ with the observation density $g_{B}$. Finally, the likelihood of data is given by
\begin{equation} \label{eq: likelihood of Y}
p_{\theta}(y_{1:T}) = \mathbb{E}_{\psi} \left[ \prod_{t = 1}^{T} g_{B}(y_{t} | X_{t}) \right].
\end{equation}
In this paper, we treat $\theta$ as unknown and seek for the MLE solution $\theta^{\ast}$ for it, which satisfies
\begin{equation} \label{eq: MLE}
\theta^{\ast} = \arg \max_{\theta \in \Theta} p_{\theta}(y_{1:T}).
\end{equation}

\subsection{Relation to the classical NMF} \label{sec: Relation to the classical NMF problem}
In the classical NMF formulation \citep{Lee_and_Seung_1999, Lee_and_Seung_2000}, given a $M \times T$ nonnegative matrix $Y = \left[ y_{1} \ldots y_{T} \right]$, we want to factorize it to $M \times K$ and $K \times T$ nonnegative matrices $B$ and $X = \left[ X_{1} \ldots X_{T} \right]$ such that the difference between $Y$ and $BX$ is minimised according to a divergence
\begin{equation} \label{eq: classic NMF solution}
(B^{\ast}, X^{\ast}) = \arg \min_{B, X} D(Y || BX).
\end{equation}
One particular choice for $D$ is the generalised Kullback-Leibler (KL) divergence which is written as
\begin{equation} \label{eq: KL divergence}
D(Y || U) = \sum_{m = 1}^{M} \sum_{t = 1}^{T} Y(m, t) \log \frac{Y(m, t)}{U(m, t)} - Y(m, t) + U(m, t) \nonumber
\end{equation}
Noticing the similarity between the generalised KL divergence and the Poisson distribution, \citep{Lee_and_Seung_1999} showed that the minimisation problem can be formulated in a MLE sense. More explicitly, the solution to
\begin{eqnarray} \label{eq: MLE for classical NMF}
& (B^{\ast}, X^{\ast}) = \arg \max_{B, X} \ell(  y_{1}, \ldots, y_{T} | B, X), & \nonumber \\
& \ell(  y_{1}, \ldots, y_{T} | B, X) = \prod_{t = 1}^{T} g_{B}\left(y_{t} | X_{t} \right) &
\end{eqnarray}
is the same as the solution to \eqref{eq: classic NMF solution}. In our formulation of the NMF problem, $X = \left[ X_{1} \ldots X_{T} \right]$ is not a static parameter but it is a random matrix whose columns constitute a Markov process. Therefore, the formulation for MLE in our case changes to maximising the expected value of the likelihood in \eqref{eq: MLE for classical NMF} over the parameter $\theta = \left( B, \psi \right)$ with respect to (w.r.t.) the law of $X$
\begin{equation} \label{eq: MLE rewritten}
\left( B^{\ast}, \psi^{\ast} \right) = \arg \max_{(B, \psi) \in \Theta } \mathbb{E}_{\psi} \left[ \ell( y_{1}, \ldots, y_{T} | B, X) \right].
\end{equation}
It is obvious that \eqref{eq: MLE} and \eqref{eq: MLE rewritten} are equivalent. We will see in Section \ref{sec: EM algorithms for NMF} that the introduction of the additional process $\{ Z_{t} \}_{t \geq 1}$ is necessary to perform MLE using the EM algorithm (see \citet{Lee_and_Seung_2000} for its first use for the problem stated in \eqref{eq: classic NMF solution}).

\section{EM algorithms for NMF} \label{sec: EM algorithms for NMF}
Our objective is to estimate the unknown $\theta$ given $Y_{1:T} = y_{1:T}$. The EM algorithm can be used to find the MLE for $\theta$. We first introduce the batch EM algorithm and then explain how an online EM version can be obtained.

\subsection{Batch EM} \label{sec: Batch EM}
With the EM algorithm, given the observation sequence $y_{1:T}$ we increase the likelihood $p_{\theta}(y_{1:T})$ in \eqref{eq: likelihood of Y} iteratively until we reach a maximal point on the surface of the likelihood. The algorithm is as follows:

Choose $\theta^{(0)}$ for initialisation. At iteration $j = 0, 1, \ldots$
\begin{itemize}
\item \textbf{E-step:} Calculate the intermediate function which is the expectation of the log joint distribution of $\left( X_{1:T}, Z_{1:T}, Y_{1:T} \right)$ with respect to the law of $\left(X_{1:T}, Z_{1:T} \right)$ given $Y_{1:T} = y_{1:T}$.
\begin{equation} \label{eq: Q function of EM}
Q(\theta^{(j)}; \theta) = \mathbb{E}_{\theta^{(j)}} \left[ \left. \log p_{\theta} ( X_{1:T}, Z_{1:T}, Y_{1:T} )  \right\vert Y_{1:T} = y_{1:T}) \right] \nonumber
\end{equation}
\item \textbf{M-step:} The new estimate is the maximiser of the intermediate function
\begin{equation} \label{eq: M step of EM}
\theta^{(j +1)} = \arg \max_{\theta} Q(\theta^{(j)}; \theta) \nonumber
\end{equation}
\end{itemize}
With a slight modification of the update rules found in \citet[Section 2]{cemgil09}, one can show that for NMF models the update rule for $B$ reduces to calculating the expectations
\begin{eqnarray} \label{eq: expectations of sufficient statistics for B}
\widehat{S}_{1, T} &=& \mathbb{E}_{\theta^{(j)}} \left[ \left. \sum_{t = 1}^{T} X_{t} \right\vert Y_{1:T} = y_{1:T} \right], \nonumber \\
 \widehat{S}_{2, T} &=& \mathbb{E}_{\theta^{(j)}} \left[ \left. \sum_{t = 1}^{T} Z_{t} \right\vert Y_{1:T} = y_{1:T} \right]  \nonumber
\end{eqnarray}
and updating the parameter estimate for $B$ as
\begin{eqnarray} \label{eq: maximisation rule for B}
B^{(j+1)} = \widehat{S}_{2, T} / \left( \mathbf{1}_{M} \left[ \widehat{S}_{1, T} \right] ^{T} \right). \nonumber
\end{eqnarray}
Moreover, if the transition density $f_{\psi}$ belongs to an exponential family, the update rule for $\psi$ becomes calculating the expectation of a $J \times 1$ vector valued function
\begin{equation} \label{eq: expectation of sufficient statistics for parameters of X}
\widehat{S}_{3, T} = \mathbb{E}_{\theta^{(j)}} \left[ \left. \sum_{t = 1}^{T} s_{3, t}( X_{t-1}, X_{t}) \right\vert Y_{1:T} = y_{1:T} \right] \nonumber
\end{equation}
and updating the estimate for $\psi$ using a maximisation rule
\begin{equation} \label{eq: maximisation rule for parameters of X}
\Lambda: \mathbb{R}^{J} \rightarrow \Psi, \quad \psi^{(j + 1)} = \Lambda \left(\widehat{S}_{3, T} \right). \nonumber
\end{equation}
Note that $s_{3, t}$ and $\Lambda$ depend on the NMF model, particularly to the probability laws in \eqref{eq: law of X} defining the Markov chain for $\{ X_{t} \}_{t \geq 1}$. Therefore, we have to find the mean estimates of the following sufficient statistics  at time $t$.
\begin{eqnarray} \label{eq: sufficient statistics in additive form}
S_{1, t}(x_{1:t}) = \sum_{i = 1}^{t} x_{i}, \quad S_{2, t}(z_{1:t}) = \sum_{i = 1}^{t} z_{i}, \nonumber \\
S_{3, t}(x_{1:t}) = \sum_{i = 1}^{t} s_{3, t}(x_{t-1}, x_{t}).
\end{eqnarray}
Writing the sufficient statistics in additive forms as in \eqref{eq: sufficient statistics in additive form} enables us to use a forward recursion to find the expectations of the sufficient statistics in an online manner. This leads to an online version of the EM algorithm as we shall see in the following section.

\subsection{Online EM} \label{sec: Online EM}
To explain the methodology in a general sense, assume that we want to calculate the expectations $\widehat{S}_{t} = \mathbb{E}_{\theta} \left[ \left.  S_{t}(X_{1:t}, Z_{1:t}) \right\vert Y_{1:t} = y_{1:t} \right]$ of sufficient statistics of the additive form
\begin{equation} \label{eq: additive form in general}
S_{t}(x_{1:t}, z_{1:t}) = \sum_{i = 1}^{t} s_{i}(x_{i-1}, z_{i-1}, x_{i}, z_{i})
\end{equation}
w.r.t.\ the posterior density $p_{\theta}(x_{1:t}, z_{1:t} | y_{1:t})$ for a given parameter value $B$. Letting $u_{t} = (x_{t}, z_{t})$ for simplicity, we define the intermediate function
\begin{eqnarray} \label{eq: forward recursion for NMF}
T_{t}(u_{t})  =  \int S_{t}(u_{1:t}) p_{\theta}(u_{1:t-1} | y_{1:t-1}, u_{t}) d u_{1:t-1}. \nonumber
\end{eqnarray}
One can show that we have the forward recursion  \citep{Del_Moral_et_al_2009, Cappe_2011}
\begin{eqnarray}\label{eq: recursion for the key function}
T_{t}(u_{t}) &=& \int \left(T_{t-1}(u_{t-1}) + s_{t}(u_{t-1}, u_{t}) \right) \nonumber \\
&& \quad\quad \quad \quad  \quad \quad \times p_{\theta}(u_{t-1} | y_{1:t-1}, u_{t}) d u_{t-1} 
\end{eqnarray}
with the convention $T_{0}(u) = 0$. Hence, $T_{t}$ can be computed online, so are the estimates
\begin{equation} \label{eq: expectation using key function}
\widehat{S}_{t}  = \int T_{t}(u_{t}) p_{\theta}(u_{t} | y_{1:t}) d u_{t}. \nonumber
\end{equation}
We can decompose the backward transition density $p_{\theta}(u_{t-1} | y_{1:t-1}, u_{t})$  and the filtering density $p_{\theta}(u_{t} | y_{1:t})$ as
\begin{align} \label{eq: backward transition kernel}
p_{\theta}(x_{t-1}, z_{t-1} | y_{1:t-1}, x_{t}, z_{t}) &= \pi_{B}(z_{t-1}|x_{t-1},y_{t-1}) \nonumber \\
& \quad \times p_{\theta}(x_{t-1} | x_{t}, y_{1:t-1}), \\
p_{\theta}(x_{t}, z_{t} | y_{1:t}) & = \pi_{B}(z_{t}|x_{t},y_{t}) p_{\theta}(x_{t} | y_{1:t}) \label{eq: filtering density}
\end{align}
where $\pi_{B}$ is defined in \eqref{eq: law of Z given X and Y}. From \eqref{eq: sufficient statistics in additive form} we know that the required sufficient statistics are additive in the required form; therefore, the recursion in \eqref{eq: recursion for the key function} is possible for the NMF model. The recursion for $S_{3, t}$ depends on the choice of the transition density $f_{\psi}$; however the recursions for $S_{1, t}$ and $S_{2, t}$ are the same for any model regardless of the choice of $f_{\psi}$. For this reason, we shall have a detailed look at \eqref{eq: recursion for the key function} for the first two sufficient statistics $S_{1, t}$ and $S_{2, t}$.

For $S_{1, t}$, notice from \eqref{eq: backward transition kernel} that, $p_{\theta}(x_{t-1}, z_{t-1}| y_{1:t-1}, x_{t}, z_{t})$ does not depend on $z_{t}$. Moreover, the sufficient statistic $S_{1, t}$ is not a function of $z_{1:t}$. Therefore, $z_{t-1}$ in \eqref{eq: forward recursion for NMF} integrates out, and $T_{1, t}$ is a function of $x_{t}$ only. Hence we will write it as $T_{1, t}(x_{t})$. To sum up, we have the recursion
\begin{eqnarray} \label{eq: recursion for T_1}
T_{1, t}(x_{t}) = x_{t} + \int T_{1, t-1}(x_{t-1}) p_{\theta}(x_{t-1} | x_{t}, y_{1:t-1}) d x_{t-1}. \nonumber
\end{eqnarray}

For $S_{2, t}$, we claim that $T_{2, t}( x_{t}, z_{t}) = z_{t} + C_{t}(x_{t})$ where $C_{t}(x_{t})$ is a nonnegative $M \times K$ matrix valued function depending on $x_{t}$ but not $z_{t}$, and the recursion for $C_{t}(x_{t})$ is expressed as
\begin{eqnarray} \label{eq: recursion C}
C_{t}(x_{t}) &=&  \int \left(C_{t-1}(x_{t-1}) + \frac{ B \odot \left( y_{t-1} x_{t-1}^{T} \right) }{\left(B x_{t-1} \right) \mathbf{1}_{K}^{T} } \right)  \nonumber \\
&& \quad\quad\quad\quad\quad\quad \times p_{\theta}(x_{t-1}|x_{t}, y_{1:t-1}) d x_{t-1} \nonumber
\end{eqnarray}
This claim can be verified by induction. Start with $t = 1$. Since $T_{2, 0} = \mathbf{0}_{M \times K}$, we immediately see that $T_{2, t}(x_{1}, z_{1}) = z_{1} = z_{1} +  C_{1}(x_{1})$ where $C_{1}(x_{1})= \mathbf{0}_{M \times K}$. For general $t > 1$, assume that $T_{2, t-1}(x_{t-1}, z_{t-1}) = z_{t-1} + C_{t-1}(x_{t-1})$. Using \eqref{eq: backward transition kernel},
\begin{align}
T_{2, t}(x_{t}, z_{t}) &= z_{t} + \int \left(  z_{t-1} + C_{t-1}(x_{t-1}) \right) \pi_{B}(z_{t-1}|x_{t-1},y_{t-1}) \nonumber \\
& \quad\quad\quad\quad\quad\quad \times p_{\theta}(x_{t-1} | x_{t}, y_{1:t-1}) d x_{t-1} d z_{t-1} \nonumber
\end{align}
Now, observe that the $\left( m, k \right)$'th element of the integral $\int z_{t-1} \pi_{B}(z_{t-1}|x_{t-1},y_{t-1}) d z_{t-1} $ is $ \frac{B(m, k) y_{t-1}(m) x_{t-1}(k) }{B(m, \cdot) x_{t-1}}$. So, we can write the integral as
\begin{equation}
\int z_{t-1} \pi_{B}(z_{t-1}|x_{t-1},y_{t-1}) d z_{t-1}  = \frac{ B \odot \left( y_{t-1} x_{t-1}^{T} \right) }{\left(B x_{t-1} \right) \mathbf{1}_{K}^{T} } \nonumber
\end{equation}
So we are done. Using a similar derivation and substituting \eqref{eq: filtering density} into \eqref{eq: expectation using key function}, we can show that
\begin{eqnarray}
\widehat{S}_{2, t} & = & \int \left( C_{t}(x_{t})  +  \frac{ B \odot \left( y_{t} x_{t}^{T} \right) }{\left(B x_{t} \right) \mathbf{1}_{K}^{T} } \right) p_{\theta}(x_{t} | y_{1:t}) d x_{t}. \nonumber
\end{eqnarray}

The online EM algorithm is a variation over the batch EM where the parameter is re-estimated each time a new observation is received. In this approach running averages of the sufficient statistics are computed \citep{Elliott_et_al_2002, Mongillo_and_Deneve_2008, Cappe_2009a, Cappe_2011}, \citep[Section 3.2.]{Kantas_et_al_2009}. Specifically, let $\gamma=\{\gamma_{t}\}_{t\geq1}$, called the step-size sequence, be a positive decreasing sequence satisfying $\sum_{t \geq1}\gamma_{t}=\infty$ and $\sum_{t \geq 1}\gamma_{t}^{2} < \infty$. A common choice is $\gamma_{t}=t^{-a}$ for $0.5 < a \leq 1$. Let $\theta_{1}$ be the initial guess of $\theta^{\ast}$ before having made any observations and at time $t$, let $\theta_{1:t}$ be the sequence of parameter estimates of the online EM algorithm computed sequentially based on $y_{1:t-1}$. Letting $u_{t} = \left( x_{t}, z_{t} \right)$ again to show for the general case, when $y_{t}$ is received, online EM computes
\begin{align} \label{eq: stochastic approximation of forward smoothing_T}
T_{\gamma, t}(u_{t}) & = \int \left( \left(1-\gamma_{t}\right) T_{\gamma, t-1}(u_{t-1}) +  \gamma_{t}s_{t}(u_{t-1}, u_{t}) \right) \nonumber \\
& \quad \quad \quad \quad \quad \times p_{\theta_{1:t}}(u_{t-1} | y_{1:t-1}, u_{t}) d u_{t-1}, \\
\mathcal{S}_{t} & = \int T_{\gamma, t}(u_{t})p_{\theta_{1:t}}(u_{t}| y_{1:t}) d u_{t} \label{eq: stochastic approximation of forward smoothing_S}
\end{align}
and then applies the maximisation rule using the estimates $\mathcal{S}_{t}$. The subscript $\theta_{1:t}$ on the densities $p_{\theta_{1:t}}(u_{t-1}|y_{1:t-1}, u_{t})$ and $p_{\theta_{1:t}}(u_{t} | y_{1:t})$ indicates that these laws are being computed sequentially using the parameter $\theta_{k}$ at time $k$, $k\leq t$. (See Algorithm \ref{alg: SMC online EM for NMF model} for details.) In practice, the maximisation step is not executed until a burn-in time $t_{b}$ for added stability of the estimators as discussed in \citet{Cappe_2009a}.

The online EM algorithm can be implemented exactly for a linear Gaussian state-space model \citep{Elliott_et_al_2002} and for finite state-space HMM's. \citep{Mongillo_and_Deneve_2008, Cappe_2011}. An exact implementation is not possible for NMF models in general, therefore we now investigate SMC implementations of the online EM algorithm.

\subsection{SMC implementation of the online EM algorithm} \label{sec: SMC implementation of the online EM algorithm}
Recall that $\left\{ X_{t}, Y_{t} \right\}_{t \geq 1}$ is also a HMM with the initial and transition densities $\mu_{\psi}$ and $f_{\psi}$ in \eqref{eq: law of X}, and the observation density $g_{B}$ in \eqref{eq: law of Y given X}. Since the conditional density $\pi_{B}(z_{t} | x_{t},y_{t})$ has a close form expression, it is sufficient to have a particle approximation to only $p_{\theta}(x_{1:t} | y_{1:t})$. This approximation can be performed in an online manner using a SMC approach.  Suppose that we have the particle approximation to $p_{\theta}(x_{1:t} | y_{1:t})$ at time $t$ with $N$ particles
\begin{eqnarray} \label{eq: particle approximation to posterior of X given Y}
p_{\theta}^{N}(d x_{1:t}|y_{1:t}) = \sum_{i = 1}^{N} w_{t}^{(i)} \delta_{ x_{1:t}^{(i)}}(d x_{1:t}), \quad \sum_{i =1}^{N} w_{t}^{(i)} = 1,
\end{eqnarray}
where $x_{1:t}^{(i)} =(x_{1}^{(i)}, \ldots, x_{t}^{(i)} )$ is the $n$'th path particle with weight $w_{t}^{(i)}$ and $\delta_{x}$ is the dirac measure concentrated at $x$. The particle approximation of the filter at time $t$ can be obtained from $p_{\theta}^{N}(d x_{1:t} | y_{1:t})$ by marginalization
\[
p_{\theta}^{N}(d x_{t}|y_{1:t}) = \sum_{i = 1}^{N} w_{t}^{(i)} \delta_{ x_{t}^{(i)}}(d x_{t}). \nonumber
\]
At time $t+1$, for each $n$ we draw $x_{t+1}^{(i)}$ from a proposal density $q_{\theta}(x_{t+1} | x_{t}^{(i)})$ with a possible implicit dependency on $y_{t+1}$. We then update the weights according to the recursive rule:
\[
w_{t+1}^{(i)} \propto \frac{w_{t}^{(i)} f_{\psi}(x_{t+1}^{(i)} | x_{t}^{(i)}) g_{B}(y_{t+1} | x_{t+1}^{(i)} )}{ q_{\theta}(x_{t+1}^{(i)} | x_{t}^{(i)})}.\nonumber
\]
To avoid weight degeneracy, at each time one can resample from \eqref{eq: particle approximation to posterior of X given Y} to obtain a new collection of particles $x_{t}^{(i)}$ with weights $w_{t}^{(i)} = 1/N$, and then proceed to the time $t+1$. Alternatively, this resampling operation can be done according to a criterion which measures the weight degeneracy \citep{Doucet_et_al_2000}.
The SMC online EM algorithm for NMF models executing \eqref{eq: stochastic approximation of forward smoothing_T} and \eqref{eq: stochastic approximation of forward smoothing_S} based on the SMC approximation of $p_{\theta}(x_{1:t} | y_{1:t})$ in \eqref{eq: particle approximation to posterior of X given Y} is presented Algorithm \ref{alg: SMC online EM for NMF model}.
\begin{alg}
\label{alg: SMC online EM for NMF model}
\textbf{SMC online EM algorithm for NMF models}
\begin{itemize}
\item \textbf{E-step:}
If t = 1, initialise $\theta_{1}$; sample $\widetilde{x}_{1}^{(i)} \sim q_{\theta_{1}}(\cdot)$, and set $w_{1}^{(i)} = \frac{ \mu_{\psi_{1}}(\widetilde{x}_{1}^{(i)}) g_{B_{1}}(y_{1} | \widetilde{x}_{1}^{(i)})}{ q_{\theta_{1}}(\widetilde{x}_{1}^{(i)})}$, $\widetilde{T}_{1, 1}^{(i)}  = \widetilde{x}_{1}^{(i)} $, $\widetilde{C}_{1}^{(i)} = 0$,  $\widetilde{T}_{3, 1}^{(i)}  = s_{3, 1}(\widetilde{x}_{1}^{(i)})$, $i=1,\ldots,N$. If $t >1$,
\begin{itemize}
\item For $i=1,\ldots,N$, sample $\widetilde{x}_{t}^{(i)} \sim q_{\theta_{t}}( \cdot | x_{t-1}^{(i)})$ and compute
\[
\widetilde{T}_{1, t}^{(i)} =  (1- \gamma_{t}) T_{1, t-1}^{(i)} + \gamma_{t} \widetilde{x}_{t}^{(i)},
\]
\[
\widetilde{T}_{3, t}^{(i)} =  (1- \gamma_{t}) T_{3, t-1}^{(i)} + \gamma_{t} s_{3, t}(x_{t-1}^{(i)}, \widetilde{x}_{t}^{(i)})
\]
\[
\widetilde{C}_{t}^{(i)}=  (1- \gamma_{t}) C_{t-1}^{(i)} + (1- \gamma_{t}) \gamma_{t-1} \frac{ B_{t} \odot \left( y_{t-1} x_{t-1}^{(i)T} \right) }{\left(B_{t} x_{t-1}^{(i)} \right) \mathbf{1}_{K}^{T} },
\]
\[
\widetilde{w}_{t}^{(i)} \propto \frac{w_{t-1}^{(i)} f_{\psi_{t}}(\widetilde{x}_{t}^{(i)} | x_{t-1}^{(i)}) g_{B_{t}}(y_{t} | \widetilde{x}_{t}^{(i)} )}{ q_{\theta_{t}}(\widetilde{x}_{t}^{(i)} | x_{t-1}^{(i)})}. 
\]
\item Resample from particles $\{ (\widetilde{x}_{t}, \widetilde{T}_{1, t}, \widetilde{C}_{t}, \widetilde{T}_{3, t} )^{(i)} \}$ for $i=1,\ldots,N$ according to the weights $\{ \widetilde{w}_{t}^{(i)} \}_{i=1,\ldots,N}$ to get $\{ ( x_{t}, T_{1, t}, C_{t}, T_{3, t} )^{(i)} \}$ for $i=1,\ldots,N$ each with weight $w_{t}^{(i)} = 1/N$.
\end{itemize}

\item \textbf{M-step:} If $t < t_{b}$, set $B_{t+1} = B_{t}$. Else, calculate using the particles before resampling
\[
\mathcal{S}_{1, t}= \sum_{i = 1}^{N} \widetilde{T}_{t}^{1 (i)} \widetilde{w}_{t}^{(i)}, \quad
\]
\[
 \mathcal{S}_{2, t} = \sum_{i = 1}^{N} \left(\widetilde{C}_{t}^{(i)} + \gamma_{t} \frac{ B_{t} \odot \left( y_{t} \widetilde{x}_{t}^{(i)T} \right) }{\left(B_{t} \widetilde{x}_{t}^{(i)} \right) \mathbf{1}_{K}^{T} } \right) \widetilde{w}_{t}^{(i)}
 \]
\[
\mathcal{S}_{3, t} = \sum_{i = 1}^{N} \widetilde{T}_{t}^{3 (i)} \widetilde{w}_{t}^{(i)},
\]
update the parameter $\theta_{t+1} = \left( B_{t+1}, \psi_{t+1} \right)$, $B_{t+1}= \frac{\mathcal{S}_{2, t}} {\mathbf{1}_{M} \left[ \mathcal{S}_{1, t} \right]^{T}}$, $\psi_{t+1} = \Lambda(\mathcal{S}_{3, t})$.
\end{itemize}
\end{alg}
Algorithm \ref{alg: SMC online EM for NMF model} is a special application of the SMC online EM algorithm proposed in \cite{Cappe_2009a} for a general state-space HMM, and it only requires $\mathcal{O}(N)$ computations per time step. Alternatively, one can implement an $\mathcal{O}(N^{2})$ SMC approximation to the online EM algorithm, see \citet{Del_Moral_et_al_2009} for its merits and demerits over the current $\mathcal{O}(N)$ implementation. The $\mathcal{O}(N^{2})$ is made possible by plugging the following SMC approximation to $p_{\theta}(x_{t-1} | x_{t}, y_{1:t-1})$ into \eqref{eq: recursion for the key function}
\begin{eqnarray} \label{eq: SMC approximation to the backward transition density}
p_{\theta}^{N}(dx_{t-1} | x_{t}, y_{1:t-1}) &=& \frac{p_{\theta}^{N}(dx_{t-1} | y_{1:t-1}) f_{\psi}(x_{t} | x_{t-1})}{\int p_{\theta}^{N}(dx_{t-1} | y_{1:t-1}) f_{\psi}(x_{t} | x_{t-1})}. \nonumber
\end{eqnarray}

\section{Numerical examples} \label{sec: Numerical examples}
\subsection{Multiple basis selection model} \label{sec: Multiple basis selection model}
\begin{figure} \label{fig: multiple basis selection results}
\vspace{-0.8cm}
\includegraphics[scale = 0.60]{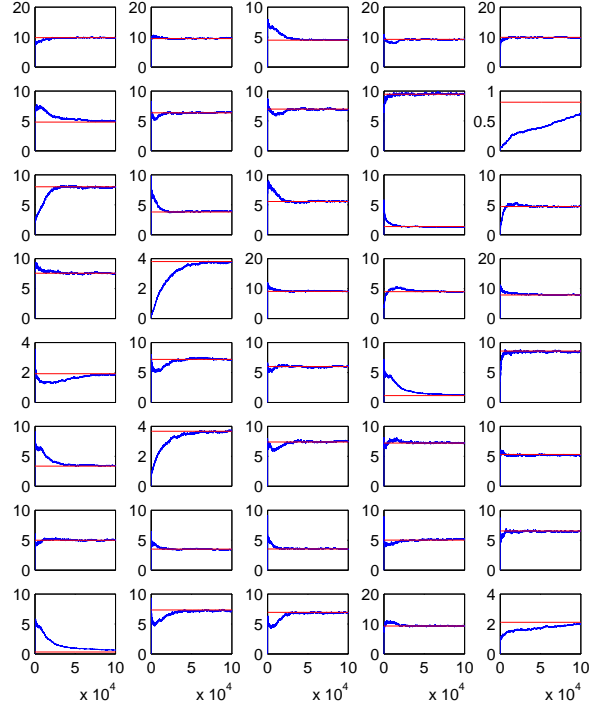}
\vspace{-1cm}
\caption{Online estimation of $B$ in the NMF model in Section \ref{sec: Multiple basis selection model} using exact implementation of online EM for NMF. The $(i, j)$'th subfigure shows the estimation result for the $B(i, j)$ (horizontal lines).}
\end{figure}

In this simple basis selection model, $X_{t} \in \{ 0, 1 \}^{K}$ determines which columns of $B$ are selected to contribute to the intensity of the Poisson distribution for observations. For $ k = 1, \ldots, K$, 
\begin{equation} \label{eq: Markov chain with 0 and 1}
X_{1}(k) \sim \mu(\cdot), \quad \text{Prob}(X_{t}(k)  = i | X_{t-1}(k) = j ) = P(j, i),  \nonumber
\end{equation}
where $\mu_{0}$ is a distribution over $\mathcal{X}$ and $P$ is such that $P(1, 1) = p$ and $P(2, 2) = q$. Estimation of $\psi = (p, q)$ can be done by calculating
\begin{align}
\widehat{S}_{3, T} &=  \mathbb{E}_{\theta} \left[ \left. \sum_{i = 1}^{T} s_{3, i}(X_{i-1}, X_{i}) \right\vert Y_{1:T} = y_{1:T} \right], \nonumber \\
s_{3, t}(x_{t}, x_{t-1}) &= \sum_{k = 1}^{K} \begin{bmatrix} I_{(0, 0)}(x_{t-1}(k), x_{t}(k)) \\ I_{0}(x_{t}(k))  \\ I_{(1, 1)}(x_{t-1}(k), x_{t}(k)) \\  I_{1}(x_{t}(k)) \end{bmatrix} \nonumber
\end{align}
and applying the maximisation rule $(p^{(j+1)}, q^{(j+1)}) = \Lambda(\widehat{S}_{3, t}^{(j)})$ where $\Lambda(\cdot)$ for this model is defined as
\[
\Lambda (\widehat{S}_{3, t} ) = ( \widehat{S}_{3, t} (1) / \widehat{S}_{3, t} (2),  \widehat{S}_{3, t} (3)/\widehat{S}_{3, t} (4) ). 
\]
Figure \ref{fig: multiple basis selection results} shows the estimation results of the exact implementation of online EM (with $\gamma_{t} = t^{-0.8}$ and $t_{b} = 100$) for the $8 \times 5$ matrix $B$ (assuming $(p, q)$ known) given the $8 \times 100000$ matrix $Y$ which is simulated $p = 0.8571, q =  0.6926$.

\subsection{A relaxation of the multiple basis selection model} \label{sec: A relaxation of the multiple basis selection model}
\begin{figure} \label{fig: example for X}
\includegraphics[height = 3cm, width = 8cm]{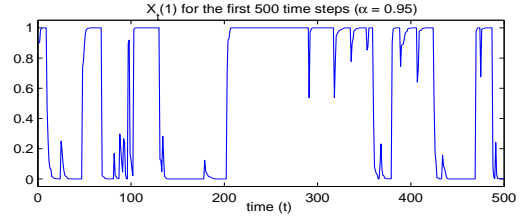}
\vspace{-0.3cm}
\caption{A realisation of $\{ X_{t}(1) \}_{t \geq 1}$ for $\alpha = 0.95$.}
\end{figure}

In this model, the process $\{ X_{t} \in (0, 1) \}_{t \geq 1}$ is not a discrete one, but it is a Markov process on the unit interval $(0, 1)$. The law of the Markov chain for $\{ X_{t} \}_{t \geq 1}$ is as follows: for $k = 1, \ldots, K$, $X_{1}(k) \sim \mathcal{U}(0, 1)$, and
\begin{align}
X_{t+1}(k) | (X_{t}(k) = x) &\sim \rho(x) \mathcal{U}(0, x) + (1 - \rho(x)) \mathcal{U}(x, 1), \nonumber \\
\rho(x)  &= \begin{cases} 
\alpha, & \text{ if } x \leq 0.5 \\
1 - \alpha, & \text{ if } x > 0.5.
\end{cases} \nonumber
\end{align}
When $\alpha$ is close to $1$, the process will spend most of its time around $0$ and $1$ with a strong correlation. (Figure \ref{fig: example for X} shows a realisation of $\{ X_{t}(1) \}_{t \geq 1}$ for $500$ time steps when $\alpha = 0.95$.)  For estimation of $\alpha$, one needs to calculate
\begin{align}
\widehat{S}_{3, T} &=  \mathbb{E}_{\theta} \left[ \left. \sum_{i = 1}^{T} s_{3, i}(X_{i-1}, X_{i}) \right\vert Y_{1:T} = y_{1:T} \right], \nonumber \\
s_{3, t}(x_{t-1}, x_{t}) &=\begin{bmatrix}  I_{A_{x_{t-1}(k)}}(x_{t-1}(k), x_{t}(k)) \\  I_{(0, 1) \times (0, 1) / A_{x_{t-1}(k)}}(x_{t-1}(k), x_{t}(k)) \end{bmatrix} \nonumber
\end{align}
where, for $u \in (0, 1)$, we define the set 
\[
A_{u} = \left( (0, 0.5] \times (0, u] \right) \cup \left( (0.5, 1) \times (u, 1) \right). 
\]
The  maximisation step for $\alpha$ is characterised as
\begin{equation} \label{eq: maximisation rule for log gaussian model}
\Lambda (\widehat{S}_{3, t} ) = \widehat{S}_{3, t} (1)/ \left( \widehat{S}_{3, t} (1) + \widehat{S}_{3, t} (2) \right). \nonumber
\end{equation}

\begin{figure} \label{fig: estimation of B}
\vspace{-0.80cm}
\includegraphics[scale = 0.60]{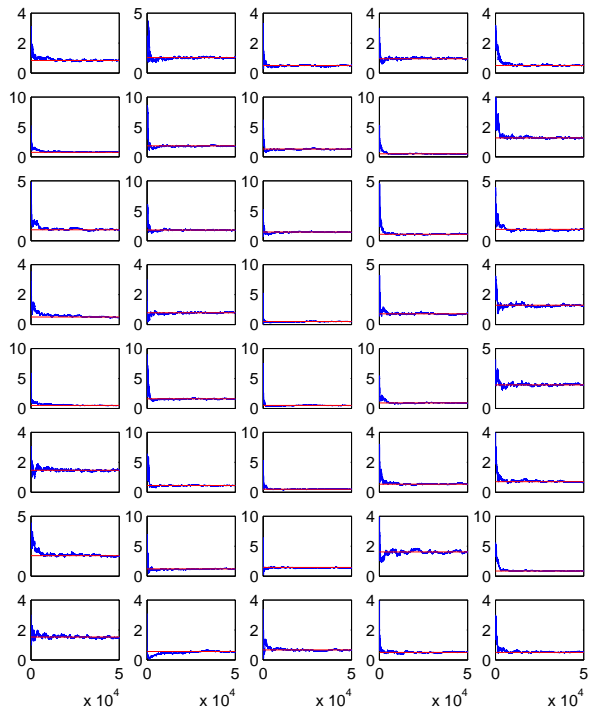}
\vspace{-1cm}
\caption{Online estimation of $B$ in the NMF model in Section \ref{sec: A relaxation of the multiple basis selection model} using Algorithm \ref{alg: SMC online EM for NMF model}. The $(i, j)$'th subfigure shows the estimation result for $B(i, j)$ (horizontal lines).}
\end{figure}

We generated a $8 \times 50000$ observation matrix $Y$ by using a $8 \times 5$ matrix $B$ and $\alpha = 0.95$. We used the SMC EM algorithm described in Algorithm \ref{alg: SMC online EM for NMF model} to estimate $B$  (assuming $\alpha$ known), with $N = 1000$ particles, $q_{\theta}(x_{t} | x_{t-1}) = f_{\varphi}(x_{t} | x_{t-1})$, $\gamma_{t} = t^{-0.8}$, and $t_{b} = 100$. Figure \ref{fig: estimation of B} shows the estimation results.

\section{Discussion}
In this paper, we presented and online EM algorithm for NMF models with Poisson observations. We demonstrated an exact implementation and the SMC implementation of the online EM method on two separate NMF models. However, the method is applicable to any NMF model where the columns of the matrix $X$ can be represented as a stationary Markov process, e.g. the log-Gaussian process. 

The results in Section \ref{sec: Numerical examples} do not reflect on the generality of the method, i.e., only $B$ is estimated but the parameter $\varphi$ is assumed to be known, although we formulated the estimation rules for all of the parameters in $\theta$. Also, we perform experiments where the dimension of the $B$ matrix may be too small for realistic scenarios. Note that in Algorithm \ref{alg: SMC online EM for NMF model} we used the bootstrap particle filter, which is the simplest SMC implementation. The SMC implementation may be improved devising sophisticated particle filters, (e.g. those involving better proposal densities that learn from the current observation, SMC samplers, etc.), and we believe that only with that improvement the method can handle more complete problems with higher dimensions.

\bibliographystyle{apalike}
\bibliography{onlinenmf,myrefs_paper}

\end{document}